# VAP: The Vulnerability-Adaptive Protection Paradigm Toward Reliable Autonomous Machines

Zishen Wan[1*], Yiming Gan[2*], Bo Yu[3], Shaoshan Liu[3], Arijit Raychowdhury[1], Yuhao Zhu[2]

[1]*Georgia Institute of Technology*  [2]*University of Rochester*  [3]*Shenzhen Institute of AI and Robotics for Society*

## 1 INTRODUCTION

The next ubiquitous computing platform, following personal computers and smartphones, is likely to be inherently autonomous in nature, encompassing drones, robots, and self-driving cars, which have moved from mere concepts in labs to permeating almost every aspect of our society, such as transportation, delivery, manufacturing, home service, and agriculture [23, 35]. By the end of 2025, it is projected that 284 million vehicles will be operating in the United States, with most of them having certain degrees of autonomy [29]. Similarly, the automated drone market is expected to reach 42.8 billion US dollars, with annual sales exceeding 2 million units [28].

Behind the proliferation of autonomous machines lies a critical need to ensure reliability [10, 34]. Recent high-profile tragedies [18] underscored the importance of building reliable computing systems for autonomous machines. For instance, almost every vendor, be it in the software, hardware, or systems segment, must conform to functional safety standards when delivering products for automotive use.

Today's resiliency solutions to autonomous machines make fundamental trade-offs between resiliency and cost, which manifests as high overhead in performance, energy consumption, and chip area. For instance, hardware modular redundancy provides high safety but more than doubles the area and energy cost of the system [1]. The reason is that today's solutions are of a "one-size-fits-all" nature: they use the same protection scheme throughout the entire software computing stack of autonomous machines.

The insight of this paper is that for a resiliency solution to provide high protection coverage while introducing little cost, we must exploit the *inherent* robustness variations in the autonomous machine software stack. In particular, we show that the different nodes in the complex software stack differ significantly in their inherent robustness toward hardware faults. We find that the front-end of an autonomous machine software stack is generally more robust, while the back-end is less so.

Building on top of the inherent differences in robustness, we advocate for a *Vulnerability-Adaptive Protection (VAP)* design paradigm. In this paradigm, the protection budget, be it spatially (e.g., modular redundancy) or temporally (e.g., re-execution), should be inversely proportional to the inherent robustness of a task or algorithm in the autonomous machine system. Designers should dedicate more protection budget to less robust tasks.

The VAP design paradigm is in stark contrast to the existing "one-size-fits-all" resiliency strategy, which uniformly applies the same protection strength to all tasks in the computing systems. As a result, existing strategies must accommodate the worst case (i.e., the least robust component), leading to a high protection overhead. In contrast, VAP wisely allocates the protection budget based on the inherent robustness of each node, achieving the same protection coverage with minimal overhead.

*Equal contributions.

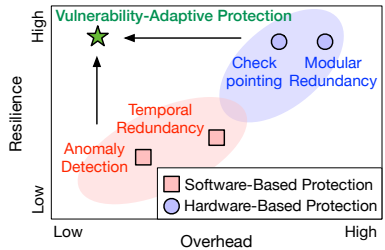

Fig. 1: Design landscape of different software and hardware-based protection techniques for resilient autonomous machines. Our proposed *Vulnerability-Adaptive Protection (VAP)* design paradigm co-optimizes performance, energy efficiency, and resilience.

In summary, we make the following contributions:

- We present a comprehensive review of the design landscape for resilient autonomous machines. We show that existing techniques are of a "one-size-fits-all" nature, where the same protection scheme is applied to the entire software stack, leading to either high overhead or low protection strength.
- We provide a thorough characterization of the inherent resilience of different tasks in widely-used, open-source software stacks for autonomous vehicles (AutoWare) and drones (MAVBench). We show that different tasks vary significantly in their resilience to hardware faults. In particular, front-end machine vision tasks that operate on massive visual data are much more resilient to faults than back-end tasks, such as planning and control, which operate on smaller data but are more sensitive to faults.
- We propose the *Vulnerability-Adaptive Protection (VAP)* paradigm for resilient autonomous machines. In VAP, we spend less protection efforts on front-end machine vision tasks and more budget on back-end planning and control tasks. Experimentally, we show that the VAP mechanism provides high protection coverage while maintaining low protection overhead on both autonomous vehicle and drone systems.

## 2 DESIGN SPACE OF RESILIENT AUTONOMOUS MACHINES

The design space of resilient autonomous machines is extremely complicated. From a system designer's perspective, latency, energy, cost, and resilience are all critical metrics that need to be cared about. We first describe the sources of errors in Sec. 2.1 and then propose the design space in Sec. 2.2. We summarize the landscape of resiliency solutions in Sec. 2.3.

### 2.1 Fault Sources

Different sources of errors can affect the resilience of autonomous machines, including adversarial attacks, software bugs, and common bit flips [33, 38, 40]. In this paper, we consider hardware bit flips that occur in a single compute cycle. This type of fault is usually referred to as soft error, serving as one of the most dominant errors influencing autonomous machine systems. The exacerbating impact of soft errors has been recently emphasized by industrial studies [13], where radiations and temperature change can result in random bit flips in silicon flip-flop units and memory cells. Future trends of increasing code complexity and shrinking feature sizes will only exacerbate the problem.

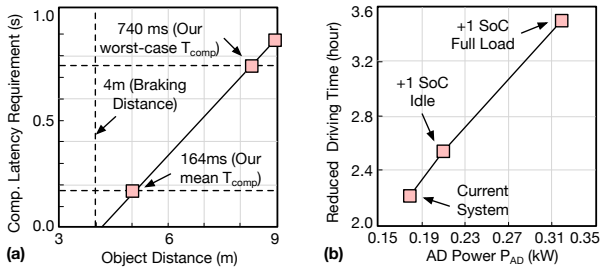

Fig. 2: The impacts of compute latency and power on autonomous vehicle systems. (a) The latency requirement becomes tighter when the object to be avoided is closer. (b) Driving time reduces as the power of the autonomous driving system increases ($P_{AD}$).

Table 1: Comparison of different hardware and software-based fault protection schemes and our proposed vulnerability-proportional protection (VAP) design paradigm. VAP exhibits high error mitigation effectiveness with low latency and energy overhead.

| Evaluation | Software (SW)-Based | | Hardware (HW)-Based | | VAP |
|---|---|---|---|---|---|
| | Anomaly Detection | Temporal Redundancy | Modular Redundancy | Checkpointing | Front-end: SW Back-end: HW |
| Latency | ✗ | ✗ | ✓ | ✗ | ✓ |
| Energy | ✓ | ✗ | ✗ | ✓ | ✓ |
| Cost | ✓ | ✓ | ✗ | ✗ | ✓ |
| Resilience | ✗ | ✗ | ✓ | ✓ | ✓ |

Soft errors can result in different misbehaviors of autonomous machine systems. The most common one is silent data corruption (SDC), where the results are different from its ground truth value [7]. In other cases, a process can hang or crash because of soft errors. In autonomous machines, the hang and crashes caused by soft errors can be solved by directly restarting the hang and crash of a process. SDCs, however, can cause more severe errors in the system. For example, an SDC on the control command can lead to an unexpected accelerating command instead of the original decelerating command, posing a severe operational safety concern.

Specifically, in our resilience analysis, we inject faults in the CPU/GPU architectural state. Memory and caches are assumed to be protected with SECDED codes. Each injected fault is characterized by its location and the injected value. The faults injected into the architectural states of these processors can manifest as errors in the inputs, outputs, and internal state of the autonomous machine modules. Bit faults cause corruption of variables when not masked in hardware and propagate to the module output.

## 2.2 Metrics and Design Constraints of Resilient Autonomous Machines

Before introducing fault protection schemes, we describe key metrics that any autonomous machine must optimize for: resilience, latency, energy consumption, and cost.

**Resilience.** A metric to quantify resilience is important in our design space as we strive to build a robust autonomous machine. We use the error propagation rate (EPR) as the resilience metric for autonomous vehicles [9]. EPR indicates the percentage of the final output of the Autonomous Vehicle (AV) software that is influenced when an error occurs at an earlier kernel. We obtain this metric by comparing the ground truth with the output after error injection. The lower the EPR, the better the resilience.

**Latency.** To guarantee resiliency, we trade off other metrics in the design space, where compute latency is the most important among them. The end-to-end compute latency, which is the time between when a new event is sensed from the surroundings and when the vehicle takes action, can impact whether an autonomous machine can stop or decelerate to avoid objects safely.

The extra computation brought by the protection scheme will increase end-to-end compute latency and further negatively impact safe object avoidance distance. Fig. 2(a) illustrates the relationship between compute latency and object avoidance distance deriving from our concrete vehicle data analysis [37]. The baseline vehicle operates at a typical speed of 5.6 $m/s$ with 4 $m/s^2$ brake deceleration, and is equipped with powerful CPU and GPU for computationally-intensive autonomy algorithms. The vehicle has an average computing latency of 164 $ms$, meaning the vehicle could avoid objects that are 5 $m$ away or farther once detected. This latency model illustrates how much compute latency matters in the end-to-end autonomous vehicle system, providing budget guidance for fault protection scheme overhead and impact analysis.

**Energy.** Energy consumption closely correlates with automotive endurance. With the trend towards electrification, the majority of vehicles in the future will be powered by batteries.

The extra energy consumed by autonomous driving computing systems will reduce the vehicle operating time and translate to revenue loss for commercial vehicles. Fig. 2(b) demonstrates the relationship between the power of the autonomous driving system and reduced driving time [37]. The vehicle is powered by batteries that have a total energy budget of 6 kilowatt hours ($kW·h$). The vehicle itself consumes 0.6 $kW$ on average, and enabling autonomous driving consumes an additional 0.175 $kW$, allowing for 7.7 hours of driving time under a single battery charge. The extra energy consumption from the protection scheme will further reduce operating time. This energy model allows us to understand how extra energy consumption brought by the protection scheme would impact the driving time and daily revenue of the vehicle.

**Cost.** Cost overhead is important to almost every vendor. We analyze extra chip area and silicon cost for autonomous vehicles and drone systems brought by various fault protection schemes.

## 2.3 Landscape of Protection Techniques

Different protection techniques exhibit distinct performance, efficiency, cost, and resilience impacts on autonomous machines. Tab. 1 compares four software and hardware protection schemes, and illustrates their tradeoff in the resilient autonomous machine design. We reveal that conventional "one-size-fits-all" approaches are limited by the tradeoff in overhead and resilience improvement.

We first analyze two software-based protection schemes. Software-based protection scheme usually exhibits advantages in lower engineering cost and power overhead, but suffer from compute latency overhead and non-completely recovery from faults.

**Anomaly detection.** Anomaly detection is generally used to identify rare observations which significantly deviate from the majority of data. In autonomous scenarios with a large number of abnormal behaviors, anomaly detection may incur high latency overhead due to the node re-execution and cannot fully mitigate fault impact due to false-positive detection in corner cases [9, 32].

**Temporal redundancy.** Temporal redundancy refers to executing the code more than once with the same piece of hardware. The redundant executions can help alleviate the threat of silent data corruption caused by soft errors as they are transient. The temporal data diversity and redundancy in the sensor data can also be exploited in detecting hardware faults [15, 16]. Temporal redundancy typically incurs large compute latency and energy overhead due to the redundant sequential executions and may not be able to detect all faults due to the continued existence of few hardware defects.

We then analyze two hardware-based fault protection schemes, namely modular redundancy and checkpointing. Hardware-based protection schemes typically offer effective error mitigation but incur significant power overheads and additional costs.

**Modular redundancy.** Modular redundancy (i.e., spatial redundancy) refers to executing the same node on two or more hardware platforms. For instance, Tesla's Full Self-Driving chip duplicates the entire processing logic, with one copy serving as a backup in case the other encounters unrecoverable errors. Similarly, Nvidia Orin chips for self-driving applications enable full system duplication and lock-step execution [1, 6]. Other established methods include triple modular redundancy, which involves three identical hardware instances with voting logic at the output. If an error affects one hardware instance, the voting logic records the majority output and masks the malfunctioning hardware. Modular redundancy is typically effective in fault detection with negligible impact on latency, though it incurs considerable energy and silicon costs.

**Checkpointing.** Checkpointing refers to periodically storing a fault-free copy of the processor state so that computation can continue from that point without altering the autonomous machine's behavior [4]. A rollback consists of a recovery mechanism that restores the processor to a previous safe state in case the autonomous machine crashes due to a failure in the underlying system. The conventional checkpointing method usually accompanies dual modular redundancy and brings large runtime overhead due to the store and retrieve procedure that may violate the real-time nature of autonomous machines (Sec. 5.2). Checkpointing can also implemented in a software manner [17] that trade-offs between efficiency and overhead. Including hardware support in saving checkpoints and re-executing can increase efficiency significantly. Since autonomous machines continuously interact with their environments and have strict real-time and efficiency requirements, we particularly refer to hardware checkpointing in the protection landscape.

Conventional "one-size-fits-all" hardware or software-based protection techniques are limited by the fundamental trade-off between performance overhead and resilience improvement in the design space of autonomous machines. Recently, ASIL decomposition [39] is proposed to decompose the automotive code with higher ASIL standards into different pieces of lower ASIL standards, and then place the decomposed pieces on different cores. As in Fig. 1, in this work, we aim to push the landscape frontier to the top-left corner with low overhead and high resilience. Thus, we leverage the insight from inherent systems performance and resilience characteristics (Sec. 3), and propose to overcome this trade-off by concurrently optimizing performance-efficiency-resilience with an intelligent *vulnerability-adaptive protection paradigm* (Sec. 4).

## 3 SYSTEM CHARACTERIZATION STUDY

This section characterizes the performance and resilience of different modules in a typical autonomous machine system. Autonomous machine computing differentiates from traditional systems in dataflow, software pipeline, compute substrate, and underlying architecture [25]. Our characterization suggests that different modules exhibit diverse performance and resilience features. The front-end of the autonomous system (sensing, localization, perception) usually has higher resilience but also higher latency and energy consumption, while the back-end (planning, decision-making, control) is more vulnerable to errors but has lower latency.

We introduce an autonomous machine system and first use the autonomous vehicle as an example. For each module, we quantify their reliability using the quantitative metric and show the inherent trade-off between resilience and performance (Sec. 3.1). We then illustrate a similar finding with drones as another example (Sec. 3.2).

### 3.1 Performance and Resilience Trade-off in Autonomous Vehicles

A typical autonomous machine system consists of five components: sensing, perception, localization, planning, and control. Sensor samples are first synchronized and processed before being used by the perception and localization modules. The localization module localizes the vehicle to the global map, and the perception module tries to understand the surroundings by detecting and tracking objects. The perception and localization results are used by the planning module to plan a path and generate control commands. The control module will smooth the control signals and transmit them through to the vehicle's Engine Control Unit. The control signals control the vehicle's actuators, such as the gas pedal, brake, and steering wheel. Each module contains one or more nodes, and each node is an individual process while the system is running.

**Front-end Modules: Sensing, Perception, and Localization.** We separate the computing pipeline of autonomous machines into front-end and back-end. The front-end consists of three modules: sensing, perception, and localization. The front-end deals with sensor data and provides semantic results for the back-end.

**Sensing.** Sensing tries to capture the environments [5, 26] and is time-consuming. We show one of the nodes in the sensing stage in Fig. 3(a) with the average latency of the specific node on the left y-axis. A node is an individual process performing certain tasks in Robot Operating System (ROS). *Ray_filter* filters the LiDAR points to represent the ground. The average runtime of *ray_filter* is 29.6 *ms*, which is a significant latency in an autonomous vehicle pipeline.

**Perception.** Perception is to build reliable and detailed representations of the dynamic surroundings based on sensory data [11, 27]. The perception module is inherently computationally intensive and usually contributes the longest latency in autonomous machines. The serial processing of detect, track, and predict in the perception pipeline exacerbates the computing latency. Although most perception nodes have been accelerated by GPUs or other accelerators, the perception stage still takes more than 100 *ms* to finish. Fig. 3(a) shows that the *vision_detect* node contributes 42.2 *ms* to the entire pipeline, and *lidar_detect* node also has a latency of 15.5 *ms*.

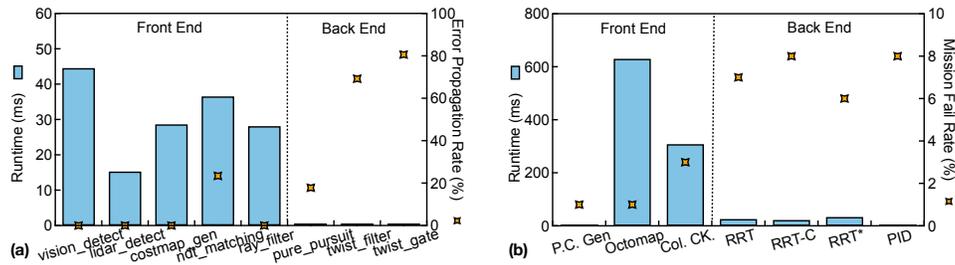

Fig. 3: The performance and resilience trade-offs of (a) autonomous vehicles (Autoware) and (b) drones (MAVBench). X-axis represents the algorithm nodes in the front-end and back-end. The left y-axis represents the measured performance of individual nodes, and the right y-axis represents resilience. The front-end exhibits high runtime and high resilience, while the back-end exhibits low runtime and low resilience.

**Localization.** Localization is to calculate the position and orientation of an autonomous machine itself in a given frame of reference. Localization algorithms usually have high computational requirements. These algorithms, such as Simultaneous localization and mapping (SLAM) [31] and Visual Inertial Odometry (VIO) [2], first capture correspondence in continuous frames, and then use multiple correspondences to solve a complex optimization problem for the final pose. A typical SLAM algorithm takes tens of *ms* latency even running on a powerful Intel CPU [22, 36]. VIO algorithm exhibits similar latency [30]. Fig. 3(a) shows that our autonomous vehicle pipeline uses *ndt_matching* as the localization algorithm, with average latency of 35.2 *ms*.

**Back-end Modules: Planning, Decision-Making, and Control.** The back-end of the autonomous machine autonomy pipeline contains three modules: planning, decision-making, and control. The results of the back-end directly control the actuators.

**Planning.** Planning tires to find a collision-free path from the current location to the destination [8, 12]. The planning algorithms usually rely on the occupancy grid that is produced by the perception module and indicates whether the locations are free or occupied. Once the occupancy grid map is generated, it will not change during one planning process. The start location on the occupancy grid map is determined by the localization module.

**Decision-Making.** Autonomous machines utilize state machines to control behavior, where the state machines is controlled by the decision-making module. For example, in Autoware, when an autonomous vehicle detects a pedestrian close by, the decision-making module will turn the vehicle's status from driving to stopping. Decision-making also relies on the results of perception and localization. Similar to the planning module, decision-making is also vulnerable to errors as it directly influences the agent's behavior.

**Control.** Control is the last module in the autonomous machine pipeline, which is responsible for smoothing the control commands. The control module is the most lightweight module in autonomous machine software but with the highest vulnerability to errors.

Fig. 3(a) shows three nodes from the backend: *pure_pursuit*, *twist_filter* and *twist_gate*. All three nodes have very low latency, which is less than 0.1 *ms*, however, they all have high EPR compared to the front-end. *pure_pursuit* has an EPR of 20.3%, *twist_filter* has an EPR of 70.8% and *twist_gate* has an EPR of 80.2%.

### 3.2 Performance and Resilience Trade-off in Autonomous Drones

In the autonomous vehicle pipeline, the front-end usually contributes to higher latency in the pipeline while being more robust. The back-end is usually low in compute complexity but is vulnerable to errors. We find a similar trend existing in autonomous drones with characterization study on MAVBench simulator [3].

Fig. 3(b) shows the trade-off between performance and resilience in drones, with the average latency and the mission failure rate. All three nodes in the front-end, *Point Cloud (P. C.) Generation*, *Octomap*, and *Collision Check (Col. CK.)*, have very high average latency but low mission failure rate. The highest mission failure rate happens at *Col. CK.*, which is only 3.6%.

We have a similar observation in the back-end. Fig. 3(b) shows that all four nodes in the back-end contribute to much less latency compared to the front-end. However, the mission failure rates of the four nodes are all significantly higher. In a typical drone autonomy pipeline evaluated on MAVBench, within the total end-to-end compute latency of 871 *ms*, the front-end modules of the drone system contribute 688 *ms* (79%), while back-end modules contribute 183 *ms* (21%). However, the average mission failure rate of the back-end is more than 2× higher compared to the front-end.

## 4 VULNERABILITY-ADAPTIVE PROTECTION DESIGN METHODOLOGY

In this section, leveraging the insights of distinct performance and resilience characteristics of front-end and back-end kernels (Sec. 3), we propose an adaptive and cost-effective protection design paradigm for autonomous machine systems, achieving high operation resilience and safety with negligible latency and energy overheads.

**Design paradigm.** The key principle of our adaptive fault protection scheme is *Vulnerability-Adaptive Protection (VAP)* - the protection budget is allocated proportionally to the inherent resilience of autonomous machine kernels. If an autonomous kernel is robust to errors, we will protect it with a lightweight method, such as software-based protection. In contrast, if the kernel is vulnerable to errors, we will spend more effort protecting it, such as hardware-based protection. This adaptive scheme can adapt to different autonomy paradigms and improve autonomous machine resilience while maintaining low computation and power overhead.

Specifically, we propose to apply software-based protection scheme on front-end kernels and hardware-based protection scheme on back-end kernels as an adaptive design paradigm, as shown in Fig. 4. This is inspired by the insights from system performance and resilience characterization on exampled autonomous vehicle and drone systems (Fig. 3), where it is well observed that the front-end kernels (e.g., sensing, perception, localization) are resilient to faults but having heavy computation, while back-end kernels (e.g., planning, decision making, control) are vulnerable to faults but have a small amount of computation.

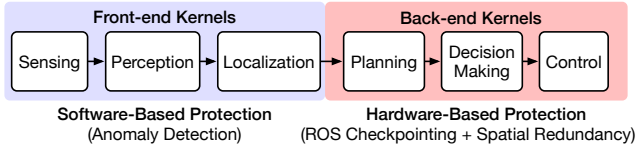

Fig. 4: Adaptive fault protection design paradigm, Vulnerability-Adaptive Protection (VAP), with software-based technique for front-end kernels and hardware-based technique for back-end kernels, based on the energy proportional protection design principle.

**Front-end - software-based protection.** We apply anomaly detection in autonomous machine front-end sensing-perception-localization kernels (Fig. 4). We leverage three insights for front-end protection. First, the vehicles or drones typically process temporal inputs and generate temporal outputs. For instance, the sequences of sensor inputs usually exhibit strong temporal consistency and continuous property. Moreover, when a vehicle is driving in a straight line, it is unlikely that the path planning module will issue a sudden actuator acceleration. Therefore, the outputs of consecutive time steps are usually bounded in the fault-free case, and errors in autonomous machines sometimes are manifested as outliers that break the temporal consistency and can be detected. Second, front-end kernels have inherent error-masking and error-attenuation capabilities through redundant information and operations such as low-pass filtering and operator union. For example, an autonomous machine can tolerate a significant input data rate drop without causing safety hazards. Third, front-end kernels exhibit rare false positive detection with anomaly detection, thus significantly reducing the node re-execution overhead and protection failure cases.

To facilitate the proposed protection scheme in a plug-and-play manner, we propose to design anomaly detection as a ROS node. In this way, the autonomous machine code can be treated as a black box, and designers can directly integrate the protection scheme in the autonomous system through standard ROS function calls.

**Back-end - hardware-based protection.** We apply modular redundancy and checkpointing in autonomous machine back-end planning-control kernels (Fig. 4), by periodically storing a fault-free copy of the architectural state and executing the same code on two hardware modules. We leverage three insights for back-end protection. First, back-end kernels are very critical to errors (Fig. 3), motivating us to strengthen fault protection with the hardware-based method. Second, the back-end nodes are extremely lightweight that do not perform any complicated computation, thus the overhead of running software calculations (e.g., anomaly detection) would be large but the overhead of hardware-based protection would be small. Third, more false positive detection cases are from the back-end in software protection, which results in potential protection failure and needs to strengthen from hardware-based protection.

To improve resilience without impacting performance, we propose a selective redundancy and ROS-based checkpointing approach. We only make redundancy copy for the hardware core running back-end modules and keep all cores running front-end modules unchanged. In ROS, we periodically queue the ROS node message during the normal process. If faults are detected, the faulty node can directly re-execute as long as the restart point is before the ROS node communication. Since all computation occurs locally before node communication and the amount of back-end computation is small, we can guarantee robustness without incurring large checkpointing overhead. This checkpointing method designed for ROS eliminates the large compute latency overhead brought by conventional architectural state checkpointing and restore methods that may violate the real-time nature of autonomous machines.

**Scalability and Adaptability.** VAP is extensible to finer-grained stage-level or node-level protection. VAP can assign suitable protection schemes to each sensing-perception-localization-planning-control stage or each ROS node given the inherent node-level robustness variations (Fig. 3). VAP is also extensible to other protection schemes, such as autoencoder-based anomaly detection and temporary redundancy [14]. Moreover, VAP is adaptive for both pre-deployment and post-deployment protection scenarios. For pre-deployment, VAP offers the methodology to characterize the vulnerability of autonomous machine pipelines thus adaptively determining the protection scheme (Sec. 3 and 4). For post-deployment, VAP can dynamically re-schedule the ROS nodes across compute cores to adaptively switch between software and hardware protection.

## 5 EVALUATION - AUTONOMOUS VEHICLE

In this section, we demonstrate the advantages of our proposed adaptive protection design paradigm in performance and resilience. Evaluated on Autoware autonomous vehicle system with design constraints (Sec. 2.2), we illustrate that the adaptive protection technique *VAP* can achieve improved resilience and lower error propagation rate (Sec. 5.1), with lower latency, energy, and system performance overhead (Sec. 5.2), compared with conventional "one-size-fits-all" software and hardware-based protection techniques.

### 5.1 Adaptive Protection Improves Resilience

The adaptive protection design paradigm *VAP* greatly reduces the error propagation rate in autonomous vehicles by leveraging the insight of high resilience of front-end kernels with inherent error-masking capabilities, and strengthening the back-end kernel resilience by hardware-based protection.

We integrate the *VAP* design into Autoware, with software-based anomaly detection in front-end modules and hardware-based modular redundancy and ROS checkpointing in back-end modules. After injecting bit-flip faults in various front-end and back-end nodes across Autoware autonomous vehicle system, we observe that the error propagation rate maintains 0%, indicating all injected faults can be masked by the inherent robustness of the system or detected and mitigated by the proposed adaptive protection scheme. This level of resilience can satisfy ASIL-D safety criteria as well.

*VAP* clearly demonstrates resilience improvement advantages with a lower error propagation rate compared with conventional software and hardware-based techniques, as illustrated in Tab. 2.

**Compared with software technique - anomaly detection.** Anomaly detection detects abnormal behaviors by leveraging the temporal consistency of autonomous machines. Evaluated on Autoware, anomaly detection can reduce the EPR from 46.5% to 24.2%.

Table 2: Comparison of our proposed adaptive protection design paradigm *VAP* with various software and hardware fault protection schemes, evaluated on end-to-end autonomous vehicle performance, energy efficiency, and resilience.

| Fault Protection Scheme | | Latency and Object Distance | | Power Consumption and Driving Time | | | | Cost | Resilience |
|---|---|---|---|---|---|---|---|---|---|
| | | Compute Latency (*ms*) | Object Avoidance Distance (*m*) | AD Component Power (*W*)* | AD Energy Change (%) | Driving Time (hour) | Revenue Loss (%) | Extra Dollar Cost | Error Propagation Rate (%) |
| **Baseline** | **No Protection** | 164 | 5.00 | 175 | – | 7.74 | – | – | 46.5 |
| **Software** | **Anomaly Detection** | 245 | 5.47 | 175 | +33.14 | 7.20 | -6.99 | negligible | 24.2 |
| | **Temporal Redundancy** | 347 | 6.05 | 175 | +75.24 | 6.62 | -14.52 | negligible | 11.7 |
| **Hardware** | **Modular Redundancy** | 164 | 5.00 | 473 | +170.29 | 5.59 | -27.78 | (CPU + GPU)×2 | 0 |
| | **Checkpointing** | 610 | 7.56 | 324 | +91.52 | 6.42 | -17.13 | (CPU + GPU)×1 | 0 |
| **Adaptive Protection Paradigm (*VAP*) Front-end Software + Back-end Hardware** | | 173 | 5.05 | 175 | +4.09 | 7.67 | -0.92 | negligible | 0 |

* The vehicle power without autonomous driving (AD) system is 600 W.

The reason that EPR is not further reduced is that in a few scenarios, the input information or output actions do have a sudden change, yet the anomaly protector treats it as an outlier and replaces it with the average value in the previous window or directly re-executes the node. These false positive errors thus will propagate and result in non-perfect results. We observe that these false positive cases mainly result from back-end modules. For example, in Autoware, 97.3% of the protection failure cases in *twist_gate* are caused by false positives, and the data is 94.5% for *twist_filter* node. This motivates us to adopt the hardware-based technique in the adaptive protection scheme to achieve improved resilience.

**Compared with software technique - temporal redundancy.** Temporal redundancy ceases fault propagation by executing the same piece of autonomous machine software code twice. Evaluated on Autoware, the temporal redundancy technique can reduce the EPR from 46.5% to 11.7%. Temporal redundancy is not guaranteed to fully mitigate hardware faults as the input to certain nodes can be faulty. Practically, instead of fully duplicating executions, the temporal redundancy scheme can trade off error detection coverage with less percentage of code duplication for lower overheads, design complexity, and availability.

**Compared with hardware technique - modular redundancy.** By executing identical software code on independent hardware pieces, the fully duplicated system is shown to be effective against soft errors on Autoware. The error propagation rate of autonomous vehicles can reduce to 0%. However, modular redundancy usually comes with a high extra dollar cost. The main overhead comes from the added silicon area and the associated non-recurring engineering, which is expected to increase as autonomous machines are increasingly integrating specialized accelerators.

**Compared with hardware technique - checkpointing.** The checkpointing scheme ceases fault propagation by retrieving the saved state, so the application is recovered from the checkpoint and continues from that point on. Combined with modular redundancy as fault detection, the checkpointing protection scheme can reduce the EPR to 0% in Autoware. However, it is to note that checkpointing and restoring procedures greatly increase compute latency that may violate the real-time nature of autonomous machines (Sec. 5.2).

> *Takeaway #1*
>
> *VAP improves resilience and reduces error propagation rate by (1) leveraging the inherent error-masking capabilities of front-end kernels and (2) strengthening back-end kernel resilience by hardware-based modular redundancy and ROS-based checkpointing scheme.*

Table 3: Compute latency breakdown analysis of different protection schemes in an end-to-end autonomous vehicle systems (unit: *ms*).

| | Perception | Localization | Planning | Control | Total |
|---|---|---|---|---|---|
| **No Protection** | 58 | 69 | 35 | 2 | 164 |
| **Anomaly Detection** | 64 | 72 | 106 | 3 | 245 |
| **Checkpointing** | 216 | 256 | 131 | 7 | 610 |
| **VAP** | 64 | 72 | 35 | 2 | 173 |

## 5.2 Adaptive Protection Reduces Performance Overhead

The proposed adaptive protection design paradigm *VAP* achieves low end-to-end latency and energy overhead by taking the advantages of (1) low cost and false-positive detection rate of software-based protection in front-end kernels and (2) low compute latency of hardware-based protection in back-end kernels.

Specifically, evaluated on Autoware autonomous vehicle systems, the proposed adaptive protection scheme slightly increases the end-to-end compute latency from 164 *ms* to 173 *ms*, resulting in a 0.05 *m* increase in object avoidance distance. The protection scheme increases the autonomous driving component energy consumption by 4.09%, resulting in a negligible 0.07 hour operation time reduction. This results from the observation that front-end nodes contribute most compute latency within the end-to-end compute pipeline, but have little to no false positive in the anomaly detection scheme, thus the overhead mainly comes from detection logic. Back-end nodes contribute little to compute latency and have light computation complexity and parameters, resulting in little modular redundancy overhead.

The adaptive protection scheme *VAP* demonstrates lower latency and energy overhead with improved end-to-end performance compared with traditional software and hardware-based techniques, as illustrated in Tab. 2 with detailed breakdown in Tab. 3.

**Compared with software technique - anomaly detection.** Anomaly detection brings performance overhead due to the detection algorithm execution and node re-compute once outliers are detected. We apply anomaly detection on multiple ROS nodes in Autoware (Fig. 3). The end-to-end compute latency increases from 164 *ms* to 245 *ms*, resulting in a longer object avoidance distance from 5 *m* to 5.47 *m* (9.4% increase). The energy consumption of the system increases by 33.14%, resulting in 0.54 *hour* driving time reduction. Since only a piece of code needs to be added to the autonomous vehicle software stack, the extra dollar cost is negligible.

**Compared with software technique - temporal redundancy.** Temporal redundancy introduces performance overhead mainly due to the redundant execution. Executing each software module twice effectively halves the performance. We apply temporal redundancy in Autoware, and the redundant execution almost doubles the compute latency from 164 *ms* to 347 *ms*. Therefore, the vehicle can only proactively plan the route to avoid obstacles at 6.05 *m* away instead of 5 *m*. The energy consumption increases by 75.24%, reducing the driving time by 1.12 hours. The temporal redundancy technique usually only involves extra software code execution with negligible engineering and silicon cost.

**Compared with hardware technique - modular redundancy.** Hardware redundancy usually trades off extra power and cost for performance. We adopt the triple modular redundancy technique as an example, and evaluate it in Autoware. By leveraging three identical main computing modules (CPU + GPU), the end-to-end compute latency is almost unchanged since modern processors usually provide hardware support to minimize the performance overhead of executing on identical hardware copies. However, due to the redundancy of hardware platform, the autonomous driving system power increases from 175 *W* to 473 *W*, reducing the driving time by 2.15 hour, which can translate to 27.78% daily revenue loss.

**Compared with hardware technique - checkpointing.** Checkpointing usually accompanies dual modular redundancy and brings a large performance overhead due to the store and retrieve procedure that may violate the real-time nature of autonomous machines. Checkpointing freezes the process and dumps the application states to the persistent storage, during which the process halts its execution without any progress. Meanwhile, checkpointing needs to be able to create globally consistent checkpoints across the entire application.

We evaluate checkpointing in Autoware, the end-to-end compute latency increases from 164 *ms* to 610 *ms*, mainly coming from the overhead of state frozen and restore. This significant latency increase results in a 51.2% longer stop distance, forcing the vehicle to avoid the obstacle 7.56 m away. That is usually not tolerable compared to their original performance. The extra checkpointing operations and redundancy increase the compute energy consumption by 91.52%, resulting in a 1.32 operation hour reduction.

Although checkpointing is considered an efficient error mitigation technique in cloud and database applications, it is not suitable for real-time applications, such as safety-critical autonomous machines, since it brings latency spikes in the critical path that may result in deadline missing. Furthermore, the checkpointing and restore times scale linearly with the memory size, thus the overhead could be bigger if the autonomy kernels take lots of memory.

> *Takeaway #2*
>
> *The adaptive protection design paradigm* VAP *reduces compute latency, energy, and end-to-end system performance overhead by taking advantage of (1) low cost and false-positive detection rate in front-end kernels and (2) low compute latency in back-end kernels of autonomous machines. Conventional "one-size-fits-all" software and hardware-based techniques are limited by fundamental tradeoffs in resilience and performance overhead.*

## 6 AUTONOMOUS DRONE VS. VEHICLE

In this section, we evaluate *VAP* on autonomous drones and focus on the difference between drone and vehicle systems. We first introduce the unique design constraints of drone systems (Sec. 6.1), and then demonstrate the resilience and performance advantages of the adaptive protection *VAP* on drone systems (Sec. 6.2).

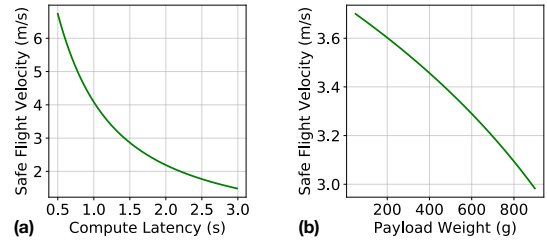

**Fig. 5: The impacts of (a) compute latency and (b) payload weight on the flight velocity and performance of autonomous drone systems.**

### 6.1 Metrics and Design Constraints of Resilient Autonomous Drones

We use resilience, latency, energy, and cost as evaluation metrics of autonomous drone systems. Different from autonomous vehicles (Sec. 2.2), drones typically have a smaller form factor, thus the extra compute latency and payload weight brought by protection schemes will impact its safe flight velocity, further impacting end-to-end system mission performance. Therefore, we focus on presenting the unique latency and energy requirements of the drone system.

**Resilience.** We use mission failure rate as the resilience metric for drones [14]. We define a failure case as the drone colliding with obstacles or failing to find a feasible path to the destination within the battery capacity limit. The reason we apply different metrics in autonomous vehicles and drones is that autonomous vehicles and drones are facing different reliability concerns. Autonomous vehicles are facing more complicated scenarios on the road. A slight difference in the command to the actuators can lead to a potential crash with other vehicles or pedestrians. Drones, however, work in scenarios with higher freedom. The chance of a crash with obstacles led by a slight trajectory change is rare.

**Latency.** The compute latency impacts the safe flight velocity of drones and needs to be short enough to ensure flight mission safety. Within the sensor rate and physics limits, as the compute latency becomes longer, the drone must lower its safe flight velocity to ensure enough time to react to obstacles without colliding.

Fig. 5(a) shows the relationship between compute latency and maximum safe flight velocity derived from analytical modeling [24] and validated in concrete real-world flight test [19, 21]. Our drone is equipped with cameras sensing the object within 4.5 *m*, and an TX2 as onboard compute to generate high-level flight commands. Our drone has an average compute latency of 871 *ms* with 2.79 *m/s* average flight velocity during an autonomous navigation task. Achieving high safe velocity is crucial as it ensures that the drone is reactive to a dynamic environment and finishes tasks quickly, thereby lowering mission time and energy [20]. This latency model allows us to understand how computing latency matters in the autonomous drone system and impact mission performance.

**Energy.** The drone is severely size, weight, and power (SWaP) constrained. A physics component change (e.g., onboard payload) will impact flight performance. Payload weight, such as redundant onboard computers and larger heatsinks, affects drone's acceleration, thus lowering its thrust-to-weight ratio and safe flight velocity.

Table 4: Comparison of our proposed adaptive protection design paradigm *VAP* with various software and hardware fault protection schemes, evaluated on end-to-end autonomous drone performance, energy efficiency, and resilience.

| Fault Protection Scheme | | Latency and Flight Time | | | Power Consumption and Flight Energy | | | | Cost | Resilience |
|---|---|---|---|---|---|---|---|---|---|---|
| | | Compute Latency (*ms*) | Avg. Flight Velocity (*m/s*) | Mission Time (*s*) | Compute Power (*W*) | Mission Energy (*kJ*) | Num. of Missions | Endurance Reduction (%) | Extra Dollar Cost | Mission Failure Rate (%) |
| Baseline | No Protection | 871 | 2.79 | 107.53 | 15 | 60.09 | 5.62 | – | – | 12.20 |
| Software | Anomaly Detection | 1201 | 2.51 | 119.52 | 15 | 66.79 | 5.05 | -10.04 | negligible | 6.44 |
| | Temporal Redundancy | 1924 | 2.14 | 140.18 | 15 | 78.34 | 4.31 | -23.30 | negligible | 3.02 |
| Hardware | Modular Redundancy | 871 | 2.74 | 109.49 | 45 | 63.13 | 5.34 | -3.79 | TX2×2 | 0 |
| | Checkpointing | 3458 | 1.75 | 171.43 | 30 | 96.76 | 3.49 | -37.90 | TX2×1 | 0 |
| Adaptive Protection Design Paradigm Frontend Software + Backend Hardware | | 897 | 2.77 | 108.30 | 15 | 60.52 | 5.58 | -0.72 | negligible | 0 |

Table 5: Compute latency breakdown analysis of different protection schemes in an end-to-end autonomous drone systems (unit: *ms*).

| | Perception | Localization | Planning | Control | Total |
|---|---|---|---|---|---|
| No Protection | 632 | 55 | 182 | 2 | 871 |
| Anomaly Detection | 645 | 60 | 493 | 3 | 1201 |
| Checkpointing | 2446 | 214 | 792 | 6 | 3485 |
| VAP | 645 | 60 | 190 | 2 | 897 |

Fig. 5(b) illustrates the relationship between drone payload weight and its maximum safe flight velocity on an exampled drone platform. As the drone gets smaller in form factor, its safe velocity would be more sensitive and affected by payload weight due to due to payload carrying capability decreasing [19]. Furthermore, flight velocity is closed correlated to the flight mission time and energy. Hence, it is essential to understand these effects when designing and evaluating fault protection schemes for drones.

> *Takeaway #3*
>
> *For small form factor autonomous machines (e.g., drones), extra compute latency and payload weight brought by fault protection schemes impact drone safe flight velocity, further impacting end-to-end system mission time, mission energy, and flight endurance.*

## 6.2 Resilience and Performance Evaluation

We evaluate the resilience and performance of the adaptive protection paradigm *VAP* on MAVBench simulator for autonomous drone systems, and demonstrate its advantages over conventional software and hardware "one-size-fits-all" techniques, as illustrated in Tab. 4 with detailed breakdown in Tab. 5.

**Adaptive protection *VAP* is cost-effective in drone system.** The proposed *VAP* exhibit high resilience on drone system with a small performance overhead. For improved operational safety, notably, the proposed adaptive protection scheme can reduce the flight mission failure rate to 0% under soft errors.

For reduced performance overhead, evaluated on a typical autonomous drone system, the adaptive protection slightly increases end-to-end compute latency from 871 *ms* to 897 *ms*. Since false positive cases are rare for frontend modules, the main overhead of anomaly detection comes from extra detection logic. This slight latency overhead results in a small drop in average flight velocity from 2.79 *m/s* to 2.77 *m/s*, consequently resulting in mission flight time slightly increasing from 107.53 *s* to 108.30 *s* and mission energy increasing from 60.09 *kJ* to 60.52 *kJ* with 0.72% less endurance.

**Conventional "one-size-fits-all" techniques bring more performance degradation in small-scale drone systems.** Conventional hardware and software protection techniques incur large compute and end-to-end performance overhead on drone systems. For example, software-based anomaly detection increases the end-to-end compute latency from 871 *ms* to 1201 *ms*, lowering the average flight velocity from 2.79 *m/s* to 2.51 *m/s*. This consequently results in 11.15% higher flight energy for the same navigation task, and the number of missions that a drone is capable of finishing is reduced from 5.62 to 5.05.

Notably, since drone has limited battery capacity and smaller form factor compared to vehicles, the compute latency overhead brings more system performance degradation. For example, software-based temporal redundancy increases the end-to-end compute latency by 2.21× in the worst scenario, which lowers the average flight velocity from 2.79 *m/s* to 2.14 *m/s*. This results in 30.37% higher flight energy and 23.30% less mission endurance. Hardware-based checkpointing increases end-to-end compute latency by 3.97×, lowering the average flight velocity to 1.75 *m/s* and resulting in 61.02% higher flight energy for the same navigation task. Similarly, extra payload weight have higher impacts on small form factor drones and result in more performance degradation.

> *Takeaway #4*
>
> *The proposed adaptive protection design paradigm VAP generalizes well to small-scale drone system with improved resilience and negligible overhead. By contrast, the large overhead from conventional "one-size-fits-all" software or hardware protection techniques results in severer performance degradation in SWaP-constrained drone systems.*

## 7 CONCLUSION AND OUTLOOK

The advent of autonomous machines has completely revolutionized several applications, and the ability of an autonomous machine to tolerate or mitigate errors is essential to ensure its functional safety. For the first time, we systematically analyze the design landscape of protection techniques for resilient autonomous machines and reveal the inherent performance-resilience trade-offs in different kernels of complex autonomous machine computing stacks. We propose an adaptive protection design paradigm, *VAP*, with front-end software and back-end hardware techniques that demonstrate cost-effectiveness in both large-scale autonomous vehicles and small-scale drone systems. We envision that the observations and design paradigms discussed in this paper will further spur a series of innovations at the algorithm, system, and hardware levels, resulting in increased deployments of intelligent swarms, enhanced autonomy, and highly efficient custom hardware designs for autonomous machine computing.

## ACKNOWLEDGEMENTS

We thank anonymous reviewers from Communications of the ACM for their valuable comments. Zishen Wan and Shaoshan Liu are the corresponding authors. The Georgia Tech authors are supported by CoCoSys, one of seven centers in JUMP 2.0, a Semiconductor Research Corporation (SRC) program.